\documentclass[letterpaper, 10 pt, conference]{ieeeconf}  

\IEEEoverridecommandlockouts                              

\usepackage{epsfig}
\usepackage{graphicx}
\usepackage{amsmath}
\usepackage{amssymb}

\usepackage{dsfont}
\usepackage{siunitx}
\usepackage{subfigure}
\usepackage{multirow}
\usepackage{booktabs} 
\usepackage{standalone}
\usepackage{xspace}
\usepackage{comment}
\usepackage{multirow}
\usepackage{xcolor}
\usepackage{pifont}
\usepackage{xfrac}
\usepackage{balance}

\usepackage[pagebackref=true,breaklinks=true,letterpaper=true,colorlinks,bookmarks=false]{hyperref}

\newcommand{\ie}[1][ ]{{\em i.\thinspace{}e\@.{}},#1}
\newcommand{\eg}[1][ ]{{\em e.\thinspace{}g\@.{}},#1}

\newcommand{\reffig}[1]{Fig.~\ref{fig:#1}}

\newcommand{\reftab}[1]{Table~\ref{tab:#1}}

\title{\LARGE \bf
UniGen: Unified Modeling of Initial Agent States and Trajectories \\for Generating Autonomous Driving Scenarios
}

\author{Reza Mahjourian$^{*}$, Rongbing Mu$^{*}$, Valerii Likhosherstov,\\Paul Mougin, Xiukun Huang, Joao Messias, Shimon Whiteson\\
\\
Waymo
\thanks{$^*$Equal contribution.}
}

\begin{document}

\maketitle
\thispagestyle{plain}
\pagestyle{plain}


\begin{abstract}
    This paper introduces UniGen, a novel approach to generating new traffic scenarios for evaluating and improving autonomous driving software through simulation.  Our approach models all driving scenario elements in a unified model: the position of new agents, their initial state, and their future motion trajectories.  By predicting the distributions of all these variables from a shared global scenario embedding, we ensure that the final generated scenario is fully conditioned on all available context in the existing scene.  Our unified modeling approach, combined with autoregressive agent injection, conditions the placement and motion trajectory of every new agent on all existing agents and their trajectories, leading to realistic scenarios with low collision rates.  Our experimental results show that UniGen outperforms prior state of the art on the Waymo Open Motion Dataset.
\end{abstract}


\section{INTRODUCTION}
	
    Autonomous Vehicles (AVs) have the potential to revolutionize transportation by providing convenient, reliable, and safe mobility for humans and goods.  However, ensuring their reliability and safety in diverse and complex traffic scenarios is a significant challenge.  To evaluate the performance of AV systems in safety-critical situations, it is necessary to capture rare or long-tail events~\cite{ding2023survey}. However, collecting a large and diverse real-world dataset of such events is difficult and expensive, due to the extensive mileage required to encounter them in the real world.  Therefore, there is a critical need for automated techniques that can generate realistic safety-critical traffic scenarios at scale.
    
    Simulation environments~\cite{dosovitskiy2017carla, lopez2018microscopic, wong2020testing} offer a solution to this problem by allowing for controlled and reproducible evaluation of AV safety and reliability.  Simulated traffic scenarios are often created with manually-designed heuristics~\cite{dosovitskiy2017carla, lopez2018microscopic}.  However, such approaches do not capture the complexity and diversity of real-world traffic scenarios.  To address these limitations, more recent approaches train deep learning models to generate traffic scenarios. Existing approaches include generating a static snapshot of the scene~\cite{tan2021scenegen, pronovost2023generating} or generating trajectories separately based on initial conditions~\cite{bergamini2021simnet, feng2022trafficgen, tan2023lctgen}. 
    
    These methods factor the scenario generation problem as
    \begin{equation}
        p(S | R) = p_\phi (S_0 | R) p_\psi (S_{1..T} | S_0, R) \label{eq:distr},
    \end{equation}
    where $R$ is the scene context including the road layout and traffic light states, $S_0$ is initial agent states, $S_{1..T}$ is agent trajectories over $T$ future timesteps, and $p_\phi$ and $p_\psi$ are probability distributions parameterized by $\phi$, $\psi$. In most prior methods, $\phi$ and $\psi$ are \textit{disjoint} and trained \textit{separately} via two different training procedures.

    However, this decomposition of the scenario generation problem is unnatural and mostly a result of incremental research advances.  Using two separate models and processes leads not only to parameter redundancy, but also an inflexible model with limited capacity for sharing the scenario context in all stages of scenario generation.

\begin{figure}
    \centering
    \includegraphics[width=\linewidth]{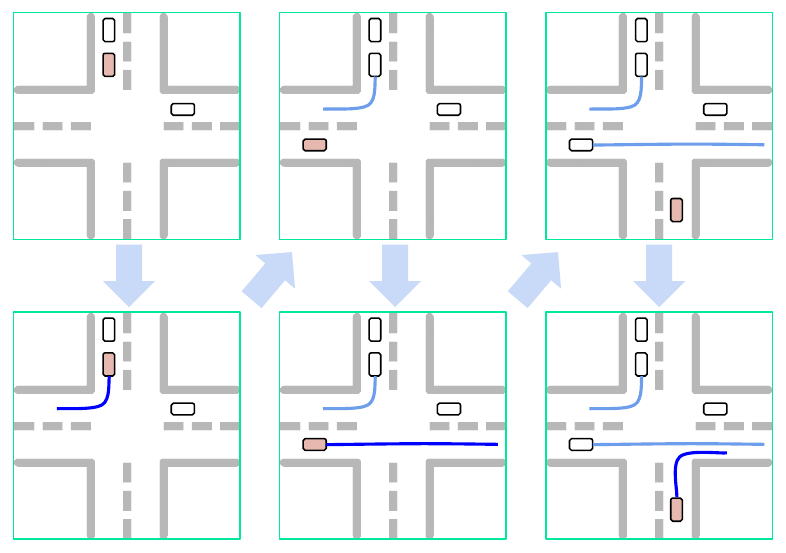}
    \caption{UniGen's autoregressive process for iteratively injecting new agents into a scenario. In each iteration, the model fully instantiates the initial state (top) and future trajectory (bottom) for a new agent (highlighted in pink).  All properties of the new agent are conditioned on the scene context and the entire trajectories for existing agents (shown in white).}
    \label{fig:teaser}
    \vspace{-1em}
\end{figure}

    In this paper, we introduce \textit{UniGen}, a multi-agent scenario generation method with state-of-the-art performance.  UniGen uses a unified model to generate the initial position, initial state attributes, and future trajectories of new agents. Using an autoregressive setup (\reffig{teaser}), UniGen can iteratively inject fully-instantiated new agents into a blank scene or a partially-populated scenario.  In particular, UniGen's model conditions all properties of the next agent on all properties of the existing agents, including agents injected in previous iterations. By ensuring consistency between initial positions and future trajectories, this approach generates more realistic traffic scenarios with lower collision rates.
    
    We evaluate UniGen on the Waymo Open Motion Dataset (WOMD)~\cite{ettinger2021large}, and show that it achieves state-of-the-art performance on both scene distribution and collision metrics.  We also report ablation experiment results to quantify the impact of each key component in UniGen's design.


\section{RELATED WORK}
\label{sec:relatedwork}

    \paragraph{Traffic scenario generation} The earliest approaches to scenario generation are procedural.  They insert traffic agents into the scene based on predefined rules and heuristics~\cite{dosovitskiy2017carla, lopez2018microscopic, kar2019meta}.  While these methods allow for manual parameter tuning to achieve reasonable agent placements and behaviors, their scalability is limited, particularly in complex urban environments with many edge cases.
    
    Building on recent advances in deep learning, several approaches model distributions over traffic scenarios based on real-world data and then draw from them.  Most learning-based scene generation methods mainly generate the initial states for the agents~\cite{tan2021scenegen, pronovost2023generating}. To fully generate a scenario, these methods rely on a separate motion forecasting model that generates trajectories given the initial agent states. SimNet~\cite{bergamini2021simnet} initializes entire scenes using a Graph Neural Network (GNN), and then uses a separate trajectory generation model to add agent motion. Similarly, TrafficGen~\cite{feng2022trafficgen} models the initial agent states using a vectorized representation~\cite{gao2020vectornet} and subsequently augments the agent data with trajectories generated by a separate model. The recently-proposed LCTGen~\cite{tan2023lctgen} model uses a shared encoder and separate transformer decoder heads to predict initial agent states and trajectories.  However, LCTGen produces the output scene all at once and, unlike our method, does not condition its predictions on the location and trajectories of previous agents, leading to less consistency between initial positions and future trajectories and higher collision rates.
    
	\paragraph{Motion forecasting} The problem of motion forecasting involves modeling the future motion of agents based on their current state and their recent motion tracks. Advances have been made through improvements in modeling inputs~\cite{gao2020vectornet, kim2022stopnet}, outputs~\cite{mahjourian2022occupancy}, agent interactions \cite{phan2020covernet, chang2019argoverse}, and multimodality~\cite{chai2019multipath, cui2019multimodal} . However, these methods are not directly applicable to scenario generation, since they assume availability of current and past state information for all agents in the scene. By contrast, our approach requires only the scene context (road layout) to create realistic proposals for injecting agents at any given $(x, y)$ location on the map and can produce future motion trajectories for agents without access to their recent motion tracks.
    
	\paragraph{Autoregressive generative models} Neural autoregressive models have found success in different domains, including text~\cite{radford2019language, oord2016wavenet}, 2D~\cite{jyothi2019layoutvae}, and 3D indoor scene generation~\cite{ritchie2019fast, wang2021sceneformer}.  For traffic scenario generation, SceneGen~\cite{tan2021scenegen} employs an autoregressive approach to insert agents into a scene one at a time.  Building on this setup, we condition new agents on both the initial states and future trajectories of existing agents---thereby increasing scenario realism and consistency while reducing collision rates.


\section{PROBLEM FORMULATION}

    We represent a traffic scenario as $(R, S)$, where $R$ represents the scene context, including road layout, and the position and state of the traffic lights, and $S = \{s^i\}, i \in \{1, \dots, N\}$ represents $N$ agents present in the scenario.  Each agent's state $s^i$ is characterized by its initial state $s^i_0$ and its future trajectory $s^i_t$ over $T$ future timesteps $t \in \{1, \dots, T\}$.  The initial agent state $s^i_0$ captures position $(x^i_t, y^i_0)$, and other attributes including width $w^i$, length $l^i$, heading angle $\theta^i_0$, and velocity $v^i_0$.  Each agent's future trajectory is captured by $\{s^i_1, \dots, s^i_T\}$.

    The task is to generate a complete driving scenario $(R, S)$, given the scene context $R$ and an initial set of agents $S_c \subset S$, which might be empty. In other words, we would like to generate the conditional distribution of all agent states $p(S | R, S_c)$.

\section{METHOD}
\label{sec:method}

\begin{figure*}[htb]
    \centering
        \includegraphics[width=\linewidth]{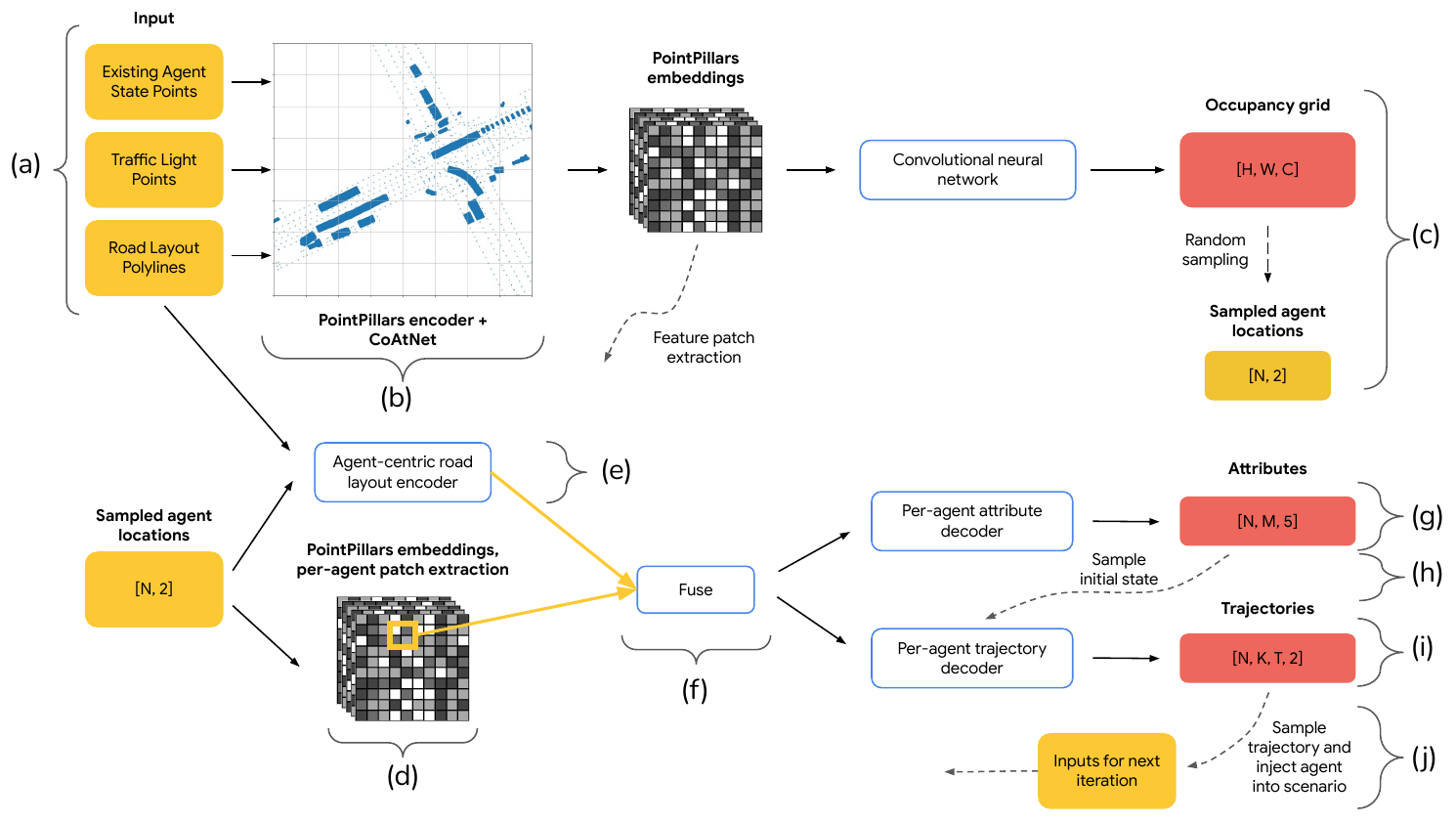}
    \caption{The overall design of UniGen. \textbf{(a)} The sparse inputs to the model consist of the polylines from the road layout, the points representing traffic lights, and the points uniformly sampled from BEV bounding boxes of existing scenario agents, if any. \textbf{(b)} The points are encoded into a dense scenario embedding. Three separate decoders predict occupancy distribution of new agents to inject, their initial states, and their future trajectories. \textbf{(c)} The occupancy decoder predicts the distribution of initial locations separately for $C$ classes of agents. In each iteration, one location is sampled from the occupancy heatmap to inject a new agent. \textbf{(d)} The location of the new agent is linearly mapped to a location in the dense scenario embedding and a feature patch is extracted surrounding that location.  \textbf{(e)} In addition, a agent-centric road layout transformer encoder extracts and encodes the road polylines normalized to the coordinate frame of the injection location. \textbf{(f)} This agent-centric road layout encoding is fused with the flattened feature patch extracted from the shared scenario embedding using a 1-layer MLP. \textbf{(g)} The product is fed to the attribute decoder to predict the initial agent states as a 5D multivariate mixture distribution with $M$ modes. \textbf{(h)} Five scalar attribute values are sampled, which together with the sampled agent location constitute the complete initial agent state.  \textbf{(i)} The trajectory decoder receives this initial agent state in addition to the fused feature encoding from (f), and predicts a set of $K$ trajectories with associated probabilities spanning over $T$ timesteps. Each trajectory waypoint is represented by a 2D Gaussian. \textbf{(j)} Finally, a single trajectory is sampled from the $K$ choices.  At this point, the new agent is fully instantiated.  The new agent is added to the scenario inputs in component (a) and the next iteration starts. \textbf{Note:} At training time, $N$ equals the number of hidden ground-truth agents.  At inference time, $N$ equals 1 for injecting a single agent in each iteration.}
    \label{fig:arch}
\end{figure*}

    \reffig{arch} illustrates the overall design of UniGen.  The model consists of a shared scenario encoder and three separate decoder heads: an occupancy predictor for generating the location of new agents, an attribute predictor for initializing agent states, and a trajectory predictor for generating future motion.  In addition to the shared whole-scene encoder which produces an embedding from all inputs, there is also a per-new-agent transformer encoder which encodes the polylines representing the road layout from the vantage point of each new agent location.  The road layout encoding produced by this encoder is passed to the agent attribute decoder and the agent trajectory decoder, in addition to the global scenario embedding.  
    

    \subsection{Masking Ground Truth Scenarios}

\begin{figure}[!t]
    \centering
        \includegraphics[width=\linewidth]{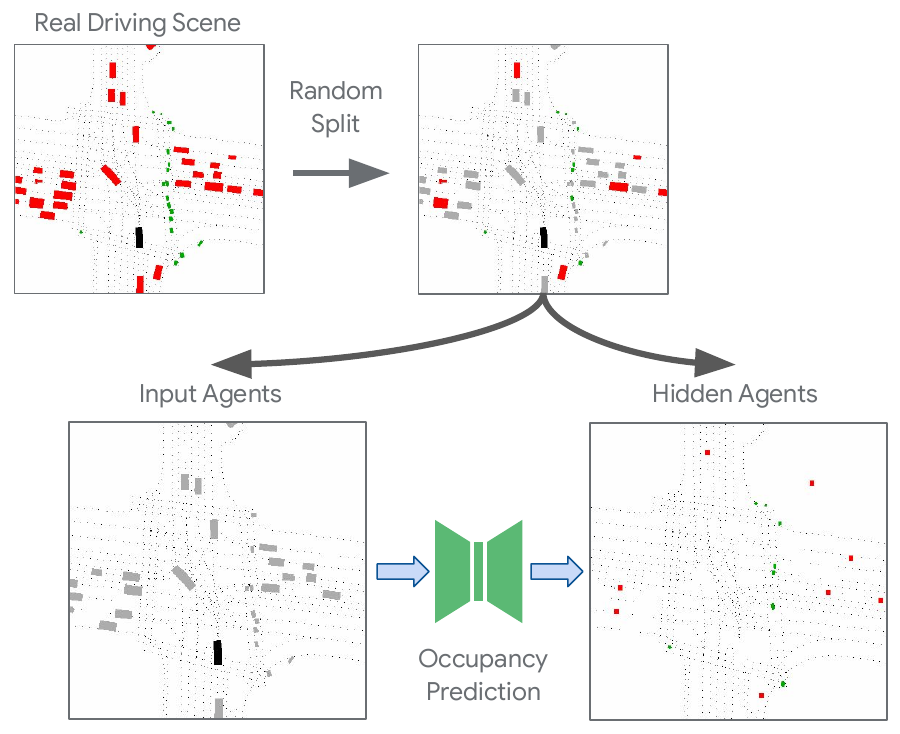}
    \caption{Masking ground-truth agents for constructing labels for training the occupancy decoder (shown), as well as the attribute and trajectory decoders (not shown).  In the ground-truth occupancy grids, only cells containing agent centers are turned on.}
    \label{fig:agent-drop}
\end{figure}

    To construct the ground-truth data for training the model, we convert every ground-truth scenario in the dataset into multiple training examples by randomly splitting the real agents into two sets of \emph{input} and \emph{hidden} agents, as illustrated in \reffig{agent-drop}. The model is trained to predict the location, attributes, and trajectories of the hidden agents, given only the input agents as inputs.  When generating each ground-truth example, a random probability $p_\text{keep}$ is first sampled to control the fraction of agents to keep in inputs.  Using a wide range of fractions allows the model to see both empty and crowded scenarios during training.
    
    This approach is conceptually similar to BERT~\cite{devlin2018bert} and masked autoencoders~\cite{he2022masked}.  The random masking encourages the model to learn the scenario dynamics in order to be able to predict the hidden agent information.  Additionally, since every ground-truth scenario can be split in many different ways, this strategy effectively increases the amount of training data available.

    \subsection{Input Representation}

    Following StopNet~\cite{kim2022stopnet}, we use sparse inputs which can adequately capture both the static scene context, and the dynamic elements of the scenario captured by the location and extents of agent bounding boxes.  More specifically, the road layout is represented by polylines that map the positions of lane centers, lane boundaries, road boundaries, crosswalks, speed bumps, and stop signs.  The state of traffic lights is also fed to the model as points placed at the end of the traffic-controlled lanes.  In addition, the model receives, as input, points uniformly sampled from the interior of Bird's-Eye View (BEV) bounding boxes for any existing or previously injected scenario agents. To encode agent trajectories, separate bounding boxes are laid out for each timestep and a separate grid of points is sampled for each timestep.  These points carry feature vectors that encode all their relevant attributes, including position, heading, velocity, and one-hot vectors that represent the timestep.  Despite being sparse, this input representation makes it easy for the model to see the regions occupied by agent bounding boxes.

    \subsection{Shared Scenario Encoder}

    All sparse inputs are encoded at once using a PointPillars encoder~\cite{lang2019pointpillars}, where each pillar encodes the points residing in it using a Multi-Layer Perceptron (MLP) and produces a single feature vector from them using max-pooling.  The dense feature map is further encoded using a CoAtNet backbone~\cite{dai2021coatnet} to encode the global interactions between the different pillar features.  The output of the shared encoder is a dense feature map with dimensions $H_d \times W_d \times D_d$.
    
    \subsection{Occupancy Prediction}
    
    The dense occupancy decoder outputs the distribution of initial positions for new agents to insert into the scenario. It receives the output of the shared encoder as input and uses a convolutional neural network to decode it into $C$ disjoint occupancy grids of size $H \times W$ corresponding to $C$ different agent classes, \eg vehicles, pedestrians, cyclists. The ground-truth occupancy grids are constructed by rendering the center points for the hidden (masked) agents as shown in \reffig{agent-drop}. Each ground-truth cell contains a binary value $O_{x, y} \in \{0, 1\}.$  The predicted occupancy values $\widehat{O}_{x, y}$ are in the range $[0, 1]$ representing the probability that a cell contains the center of some hidden agent belonging to the corresponding class.  The occupancy decoder is trained using a cross-entropy loss.  At inference time, the initial position $(x^i_0, y^i_0)$ of the new agent is determined by sampling a cell from the predicted occupancy grid.  At training time, we simply use the ground-truth positions of the hidden agents.

    \subsection{Agent-Centric Road Layout Encoder}

    While the shared scenario embedding is useful for capturing the global interactions between agents and road elements, its low spatial resolution is not ideal for regressing location-sensitive attributes like heading, which can be radically different for two nearby agents in opposing lanes.  To this end, when predicting agent attributes and trajectories, we augment the shared scenario embedding with an agent-centric road layout embedding, which is obtained as follows.  Given the sampled position $(x^i_0, y^i_0)$ for the new agent, we extract and normalize a set of road layout polylines around that position and encode them using a transformer encoder similar to Wayformer~\cite{nayakanti2022wayformer}. The encoder produces a $1 \times D_r$ feature vector per agent.

    \subsection{Per-Agent Feature Fusion}

    For each new agent, we bilinearly sample a $k \times k \times D_d$ feature patch from the shared encoder's output, where $k$ is a fixed hyperparameter.  The location of the patch in the shared embedding is determined by linearly mapping the initial agent position $(x^i_0, y^i_0)$ in the scene to a location in the dense feature map $H_d \times W_d$.  For each agent, we also obtain a $1 \times D_r$ feature vector from the agent-centric road layout encoder.  These two feature maps are flattened and passed through a 1-layer MLP to obtain a $1 \times D$ feature vector.

    \subsection{Agent Attribute Prediction}

    While the position $(x^i_0, y^i_0)$ of the new agent is sampled from the occupancy grid, its other initial state attributes are predicted by the attribute decoder.  For each agent, the additional state attributes are captured by five values: width $w^i$, length $l^i$, a two-dimensional unit heading vector $(\cos(\theta^i_0), \sin(\theta^i_0))$, and a scalar speed, which combined with the predicted heading can produce the velocity vector $v^i_0$.
    
    The attribute decoder predicts $M$ distinct modes and associated probabilities for each initial state attribute $a$.  Inspired by trajectory prediction methods~\cite{nayakanti2022wayformer}, we use a loss function consisting of classification and regression terms to learn the distribution of attributes. For each attribute $a$, let $\hat{m}^*$ denote the index of the predicted mode closest to the ground truth, and let $\text{max}(\hat{m})$ denote the index of the most likely mode according to the model.  The classification term uses cross the entropy loss to maximize the probability of selecting $\hat{m}^*$. The regression term minimizes the L1 distance between the ground-truth value $a$ and the value of the most likely mode $\hat{a}_{\text{max}(\hat{m})}$.  We found this choice to be more effective than minimizing the distance to the closest mode $\hat{a}_{\hat{m}^*}$ as done in trajectory prediction losses~\cite{nayakanti2022wayformer}.  We also scale all attribute values to have comparable loss magnitudes.
    
    \begin{equation}
    L_{a} = \underbrace{-\text{log Pr}(\hat{m}^*)}_{\text{classification loss}} + \underbrace{\left| a - \hat{a}_{\text{max}(\hat{m})} \right|}_{\text{regression loss}}.
    \end{equation}

    \subsection{Trajectory Prediction}

    We adapt the transformer-based trajectory decoder and losses from MultiPath++~\cite{varadarajan2022multipath++} and Wayformer~\cite{nayakanti2022wayformer}.  However, these trajectory prediction methods require the current and recent positions of agents, while our approach can predict trajectories given only features sampled at arbitrary locations on the map.  Our trajectory decoder takes the same inputs as the attribute decoder, namely a patch from the shared scenario embedding and the per-agent encoding of road layout polylines around the agent.  The trajectory decoder also receives, as initial state, the initial position and heading of the agent. The initial position is already determined at this point.  At evaluation and inference time, the initial heading is determined by the attributes sampled from the outputs of the attribute decoder.  At training time, we simply use the ground-truth state of the hidden agents.

    \subsection{Autoregressive Scenario Generation}

    We use an autoregressive approach to generate new scenarios at inference time by injecting new agents into the scenario one at a time.  In other words, we factor the conditional distribution for generating agent $i$ as:
    \\
    \begin{equation}
    P(s^i | R, s^1, s^2, \dots, s^{i-1}).
    \end{equation}
    
    For each new agent, an initial position $(x^i_0, y^i_0)$ is first sampled from the distribution of agent positions predicted in the occupancy grid.  This sampled position is used to extract a fused feature vector needed to predict a distribution of initial state attributes $w^i, l^i, \theta^i_0, v^i_0$.  Sampling a set of attributes from this distribution instantiates the initial state of the agent.  This initial agent state together with the fused feature vector is used to predict a distribution of future trajectories for the agent. A specific trajectory is then sampled from this distribution to fully instantiate and inject the agent, and complete one iteration of autoregressive generation.  In the next iteration, the newly-generated agent is included as part of the model inputs, influencing \emph{all} properties of the subsequent agents generated by the model.  This autoregressive approach, coupled with the unified model, yields realistic scenarios where agent initial states and future trajectories are consistent with each other.

    
    \section{EXPERIMENTAL SETUP}

    \paragraph{Dataset}
    We train our model on real traffic scenarios from the Waymo Open Motion Dataset~\cite{ettinger2021large}.  The training set contains about 69,500 scenarios and we evaluate on a subset of 1000 examples from the validation set.  Each scenario contains 8$s$ agent future trajectories recorded at 10 Hz.

    When masking agents in the ground-truth training set, we sample the per-scenario probability $p_\text{keep}$ for keeping agents in inputs uniformly from $[-0.3, 0.9]$ while clipping negative values to zero.  This ensures adequate exposure to blank scenes during training.  For the evaluation set, we remove all agents from the scenario and then use our method to inject the same number of agents into the blank scene.  The resulting scenario is then compared with the ground truth using scene and trajectory distribution metrics. 

    \paragraph{Hyperparameters}
    We model the scenes as a $120 m \times 120 m$ region centered on the AV.  We filter out agents which are outside this BEV FOV and agents which do not have valid 8$s$ future trajectories.  Ground-truth occupancy grids are created and predicted at a resolution of $H \times W = 384 \times 384$ for $C = 3$ classes of vehicles, pedestrians, and cyclists.  Each grid cell corresponds to a $31.2 cm \times 31.2 cm$ region of the world.
    
    The scenario encoder has $128 \times 128$ pillars and produces a $128 \times 128 \times 64$ output. The CoATNet backbone outputs a  $H_d \times W_d \times D_d = 32 \times 32 \times 64$ dense feature map, from which we extract feature patches of size $k \times k = 5 \times 5.$
    
    The agent-centric road layout encoder is a 4-layer transformer with 256 KV hidden size, and outputs a $1 \times 256$ feature vector.  A 1-layer MLP fuses this vector with the flattened dense feature map into a $1 \times D = 1 \times 512$ feature vector per agent.
    
    The attribute decoder is an MLP with $4$ hidden layers, each containing $1024$ hidden units with ReLU~\cite{relu} activations. The attribute decoder predicts a distribution with $M = 8$ modes.  The trajectory decoder is an 8-layer transformer with 512 KV hidden size, predicting $K = 64$ different trajectories with associated probabilities.  At inference time we sample the most likely trajectory for each agent.

	\paragraph{Baselines}
	We compare our method with TrafficGen~\cite{feng2022trafficgen} and the non-conditioned variant of LCTGen~\cite{tan2023lctgen} on scene generation and motion prediction metrics.  Unlike our method, TrafficGen is limited to vehicles only. We use the pretrained model and scenario generation code provided by the authors.  For a fair assessment of the underlying model's performance, we use TrafficGen's scenario generation code without its collision check and resampling logic.

	\paragraph{Ablations}
	To investigate the effect of different components in our approach, we train five variants of UniGen.  \emph{UniGen Joint} uses a unified model to generate all outputs but does not use the agent-centric road layout encoder or condition its predictions on future trajectories.  \emph{UniGen Separate} is similar but uses two separate models to predict initial agent states and agent trajectories, as in all prior methods.  \emph{UniGen w/ Agent-Centric R} improves upon \emph{UniGen Joint} by adding the agent-centric road layout encoder.  \emph{UniGen w/ Traj. Inputs} improves upon \emph{UniGen Joint} by processing future agent trajectories in the shared scenario encoder.  Finally, \emph{UniGen Combined} adds all these improvements.
	
	\paragraph{Initial State Metrics}
    We employ the Maximum Mean Discrepancy (MMD)~\cite{borgwardt2006integrating} metric, denoted as \(\mathrm{MMD}^2(X, Y)\), to quantify the similarity between the original and model-generated distributions for initial agent states. Given sets of samples \(X = \{x_1, ..., x_m\}\) and \(Y = \{y_1, ..., y_n\}\), and a Gaussian kernel function \(k\), an empirical estimate of \(\mathrm{MMD}^2(X, Y)\) is given by
    \begin{equation}
        \frac{1}{m^2}\sum_{i,j}k(x_i, x_j) - \frac{2}{mn}\sum_{i,j}k(x_i, y_j) + \frac{1}{n^2}\sum_{i,j}k(y_i, y_j),
    \end{equation}
    where \(k(x_i, x_j)\) denotes the kernel function evaluated at samples \(x_i\) and \(x_j\), and similarly for \(k(x_i, y_j)\) and \(k(y_i, y_j)\).
    We compute MMD separately for each initial state attribute, including position, bounding box size, heading, and velocity vector.  The MMD metric is computed for each scenario separately using a Gaussian kernel and then averaged over the validation set into a single number. We also report the Static Collision Rate (SCR), \ie the percentage of agents with overlapping initial state bounding boxes per scenario.
	
	\paragraph{Motion Metrics}
	Evaluating motion predictions for scenario generation is not straightforward since the model generally injects agents in positions that differ from the ground truth.  Some methods~\cite{feng2022trafficgen, tan2023lctgen} use matching algorithms to pair up and reorient predicted agents with ground-truth agents and then compute standard trajectory metrics such as ADE and FDE~\cite{chang2019argoverse}.  This may be suboptimal since the trajectory alignment quality is mostly affected by the initial position of the agents.  For example, consider cases where the closest ground-truth agent may be in the opposing lane.
	
	Inspired by the Waymo Sim Agents Challenge~\cite{montali2023waymo}, our motion evaluation incorporates Dynamic Collision Rate (DCR), the average percentage of agents with overlapping trajectory bounding boxes per scenario. We also report MMD metrics on motion attributes including velocity, acceleration, distance to the nearest agent, and distance to the road edge. Velocity and acceleration for each timestep are ascertained utilizing finite difference method.
	
\section{RESULTS}\label{sec:result}

\newcommand\fboxgt{\fcolorbox{green}{white}}
\newcommand\fboxsample{\fcolorbox{orange}{white}}

\begin{figure*}[h]
\centering
\begin{tabular}{cccc}
\multicolumn{1}{c}{\textbf{Ground Truth}} & \multicolumn{1}{c}{\textbf{Sample 1}} & 
\multicolumn{1}{c}{\textbf{Sample 2}} & 
\multicolumn{1}{c}{\textbf{Sample 3}}\\
\fboxgt{\includegraphics[width=0.21\linewidth]{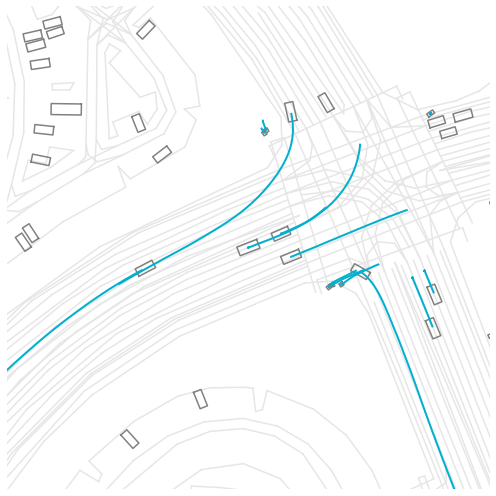}} & 
\fboxsample{\includegraphics[width=0.21\linewidth]{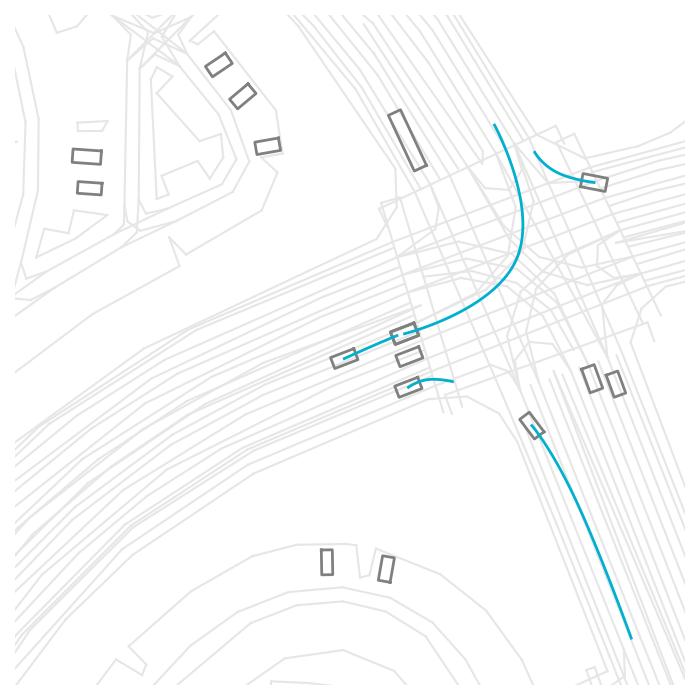}} & 
\fboxsample{\includegraphics[width=0.21\linewidth]{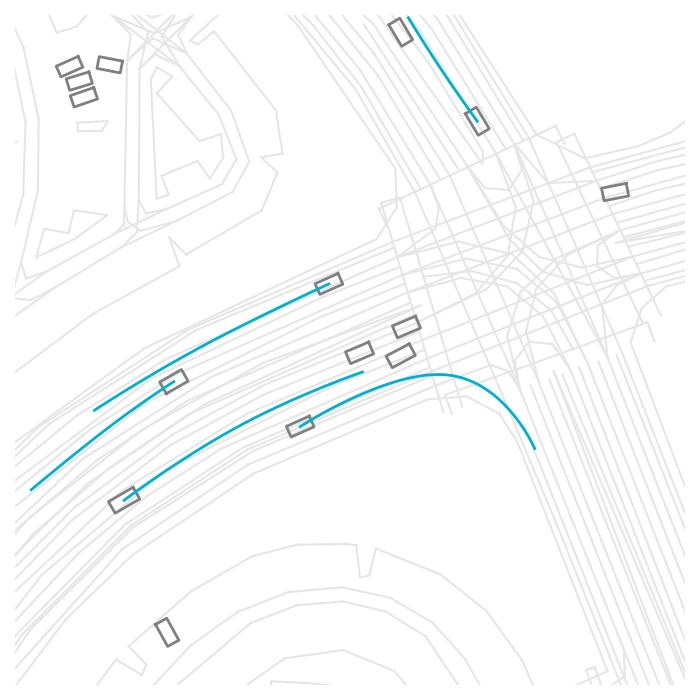}} & 
\fboxsample{\includegraphics[width=0.21\linewidth]{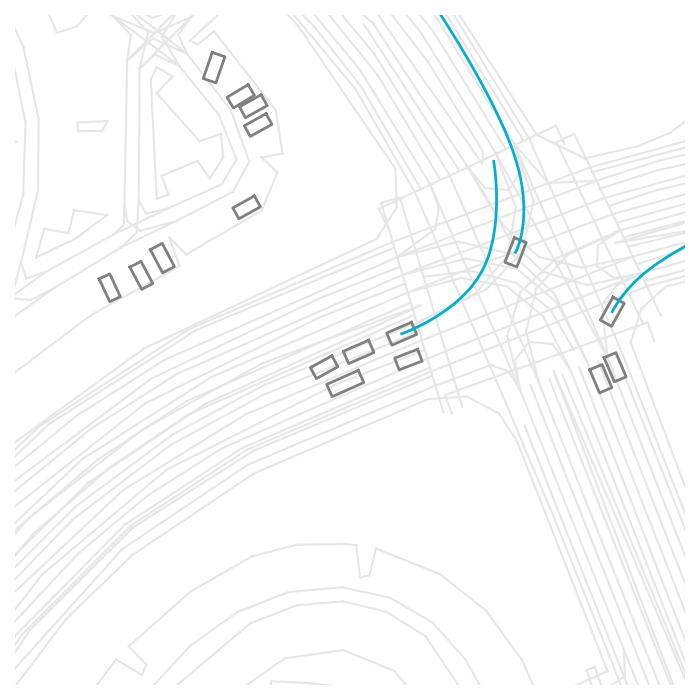}} \\
\\
\fboxgt{\includegraphics[width=0.21\linewidth]{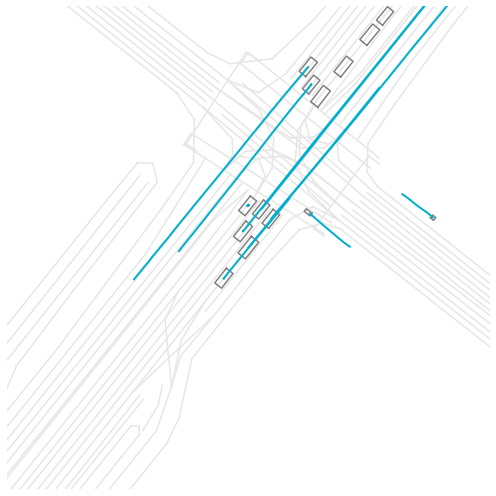}} & 
\fboxsample{\includegraphics[width=0.21\linewidth]{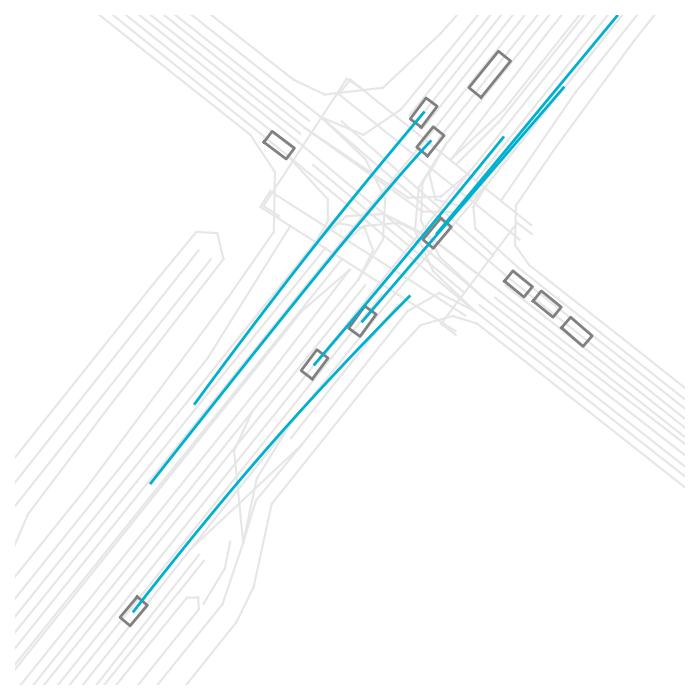}} & 
\fboxsample{\includegraphics[width=0.21\linewidth]{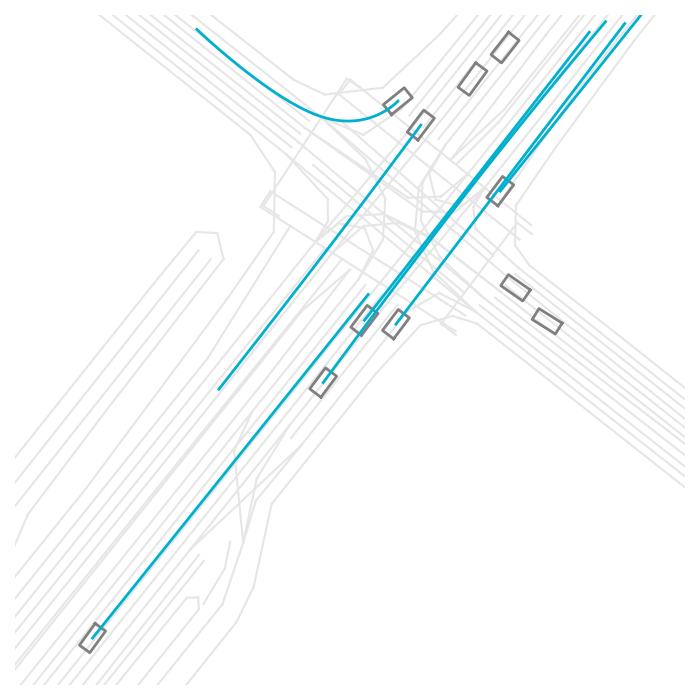}} & 
\fboxsample{\includegraphics[width=0.21\linewidth]{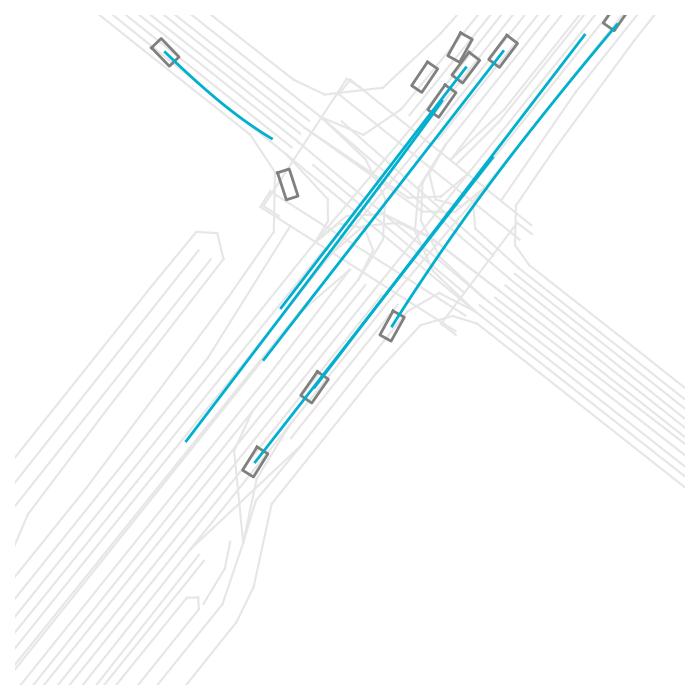}} \\
\end{tabular}
\caption{Example scenarios generated by UniGen. The left column shows two sample ground-truth scenarios. For each scenario, we remove all agents and generate new agents with trajectories given just the road layout.  We apply the method three separate times resulting in three different generated scenarios.}
\label{fig:viz}
\end{figure*}

{
\setlength{\tabcolsep}{4pt}
\renewcommand{\arraystretch}{1.3} 
\begin{table}[h]
    \caption{Initial state MMD metrics and static collision rates} 
    \vspace{-1em}
	\begin{center}
    	\resizebox{\linewidth}{!}{%
    	\begin{tabular}{@{}lccccc@{}} \toprule
    	    \textbf{Method} & SCR (\%) & Position & Heading & Size & Velocity \\
    	    \midrule
    	    Ground Truth & 0.14 & - & - & - & - \\
    	    TrafficGen~\cite{feng2022trafficgen} & 12.71 & 0.1451 & 0.1325 & 0.0926 & 0.1733 \\
            LCTGen~\cite{tan2023lctgen} & N/A & 0.1319 & 0.1418 & 0.1092 & 0.1948 \\
            UniGen Separate & 1.82 & 0.1357 & 0.2203 & 0.0835 & 0.1910 \\
            UniGen Joint & 1.87 & 0.1323 & 0.2251 & 0.0831 & 0.1915 \\
            UniGen w/ Agent-Centric R & 1.16 & 0.1217 & \textbf{0.1095} & 0.0817 & 0.1679 \\
            UniGen w/ Traj. Inputs & 1.35 & \textbf{0.1197} & 0.1897 & 0.0826 & 0.1657 \\
            UniGen Combined  & \textbf{1.13} & 0.1208 & 0.1104 & \textbf{0.0815} & \textbf{0.1591} \\
            \bottomrule
        \end{tabular}}
     \end{center}
     \vspace{-1em}
    \label{tab:scene-metrics} 
\end{table}
}

{
\setlength{\tabcolsep}{4pt}
\renewcommand{\arraystretch}{1.3} 
\begin{table}[h]
    \caption{Motion MMD metrics and 8s trajectory collision rates}
    \vspace{-1em}
	\begin{center}
    	\resizebox{\linewidth}{!}{%
    	\begin{tabular}{@{}lccccc@{}} \toprule
    	    \textbf{Method} & DCR (\%) & Speed & Acceleration & Dist. to & Dist. to\\
    	     & & & & Nearest & Road Edge\\
    	    \midrule
    	    Ground Truth & 1.20 & - & - & - & - \\
    	    TrafficGen~\cite{feng2022trafficgen} & 19.05 & 0.207 & 0.133 & 0.205 & 0.258 \\
    	    LCTGen 5s Traj~\cite{tan2023lctgen}  & 8.38 & N/A & N/A & N/A & N/A \\
    	    UniGen Separate  & 7.71 & 0.220 & 0.144 & 0.155 & 0.190  \\
    	    UniGen Joint & 7.69 & 0.223 & 0.145 &0.153 & 0.193 \\
    	    UniGen w/ Agent-Centric R & 6.72 & 0.199 & 0.105 & 0.171 & 0.194 \\
    	    UniGen w/ Traj. Inputs & 5.21 & 0.225 & 0.142 & 0.153 & 0.181 \\
    	    UniGen Combined & \textbf{4.63} & \textbf{0.186} & \textbf{0.101} & \textbf{0.136} & \textbf{0.175} \\
    	    \bottomrule
        \end{tabular}}
     \end{center}
     \vspace{-1em}
    \label{tab:motion-metrics} 
\end{table}
}
    \reftab{scene-metrics} shows MMD and static collision metrics over the initial agent states. UniGen significantly outperforms prior methods on static MMD metrics.  Ablation experiments show that including the agent-centric road layout encoder has the most impact on improving initial state metrics.  Conditioning the predictions on future trajectories for existing agents greatly improves the metrics as well.  The multitask \emph{UniGen Joint} performs similarly to \emph{UniGen Separate} with a dedicated trajectory model, while also allowing for encoding agent trajectories, which is not possible in \emph{UniGen Separate}.

    \reftab{motion-metrics} shows the MMD metrics over motion attributes and the dynamic collision metrics.  UniGen outperforms prior methods on all available metrics.  The ablation experiments show that the collision metrics improve the most when we include the future agent trajectories in the scenario encoder.  On the other hand, location- and heading-sensitive attributes like speed and acceleration improve the most by adding the agent-centric road layout encoder, which allows the model to react to the road layout around the injection position with high precision.
    
    Despite significantly improving SCR and DCR metrics compared to prior works, still some generated scenarios contain collisions.  While static collisions might be influenced by noisy bounding boxes and presence of box overlaps in ground-truth data, dynamic collisions are mainly an indication of limited capacity in the model to incorporate future trajectories of existing scenario agents, which can be addressed in future work.
    
    \reffig{viz} shows sample scenarios generated by UniGen.  The positions, initial states, and trajectories mimic the behavior of ground-truth scenario agents.  By conditioning the predictions on the entire state attributes of existing agents, UniGen increases the distributional realism of agent interactions and the consistency of the generated scenarios.


\section{CONCLUSIONS}
\label{sec:conclusion}

	This paper introduces UniGen, a novel scenario generation method with state of the art performance.  UniGen generates multimodal distributions for agent positions, initial state attributes, and future trajectories.  Using a unified model allows UniGen to fully condition all properties of every new agent on all properties of existing agents.  In particular, conditioning the predictions on trajectories of existing scenario agents greatly improves the consistency of the generated scenarios as reflected by the improved distributional metrics and lower collision rates.  Furthermore, employing an agent-centric road layout encoder greatly improves the precision of the model and its control over location- and heading-sensitive attributes.

\newpage

\balance

\bibliographystyle{IEEEtran}
\bibliography{bibliography}  


\end{document}